\documentclass[letterpaper]{article} 
\usepackage{aaai24}  
\usepackage{times}  
\usepackage{helvet}  
\usepackage{courier}  
\usepackage[hyphens]{url}  
\usepackage{graphicx} 
\usepackage{natbib}  
\usepackage{caption} 
\usepackage{amssymb}
\usepackage{amsmath}
\usepackage{amsthm}
\usepackage{color}
\usepackage{comment}
\usepackage{graphicx}
\usepackage{float}
\usepackage{subfigure}
\usepackage{booktabs}
\usepackage{multirow}
\usepackage{amsmath}
\usepackage{amssymb}
\usepackage{color}
\usepackage{multirow}
\newtheorem{problem}{Problem}
\newtheorem{definition}{Definition}
\newtheorem{theorem}{Theorem}
\newtheorem{proposition}{Proposition}
\newcommand{\methodname}{{\tt{MgCSL}}}
\usepackage{algorithm}
\usepackage{algorithmic}
\usepackage{newfloat}
\usepackage{listings}
\DeclareCaptionStyle{ruled}{labelfont=normalfont,labelsep=colon,strut=off} 
\floatstyle{ruled}
\newfloat{listing}{tb}{lst}{}
\floatname{listing}{Listing}
\title{Multi-granularity Causal Structure Learning}
\author{Paper ID: 10491}
\author{
  Jiaxuan Liang\textsuperscript{\rm 1,2},
  Jun Wang\textsuperscript{\rm 2},
    Guoxian Yu\textsuperscript{\rm 1,2}\thanks{Corresponding author.},
    Shuyin Xia\textsuperscript{\rm 3,4},
    Guoyin Wang\textsuperscript{\rm 3,4}
}
\affiliations{
    \textsuperscript{\rm 1}School of Software, Shandong University, Jinan, China\\
    \textsuperscript{\rm 2}SDU-NTU Joint Centre for AI Research, Shandong University, Jinan, China\\
     \textsuperscript{\rm 3}Chongqing Key Laboratory of Computational Intelligence, Chongqing Uni. of Posts and Telecom., Chongqing, China\\
     \textsuperscript{\rm 4}MOE Key Laboratory of Big Data Intelligent Computing, Chongqing Uni. of Posts and Telecom., Chongqing, China\\
    
    jxliang@mail.sdu.edu.cn, \{kingjun, gxyu\}@sdu.edu.cn, \{xiasy, wanggy\}@cqupt.edu.cn
}

\usepackage{bibentry}

\begin{document}

\maketitle

\begin{abstract}
Unveil, model, and comprehend the causal mechanisms underpinning natural phenomena stand as fundamental endeavors across myriad scientific disciplines. Meanwhile, new knowledge emerges when discovering causal relationships from data. Existing causal learning algorithms predominantly focus on the isolated effects of variables, overlook the intricate interplay of multiple variables and their collective behavioral patterns. Furthermore, the ubiquity of high-dimensional data exacts a substantial temporal cost for causal algorithms. In this paper, we develop a novel method called \textbf{\methodname{}} (\underline{M}ulti-\underline{g}ranularity \underline{C}ausal \underline{S}tructure \underline{L}earning), which first leverages sparse auto-encoder to explore coarse-graining strategies and causal abstractions from micro-variables to macro-ones. \methodname{} then takes multi-granularity variables as inputs to train multilayer perceptrons and to delve the causality between variables. To enhance the efficacy on high-dimensional data, \methodname{}  introduces a simplified acyclicity constraint to adeptly search the directed acyclic graph among variables. Experimental results  show that \methodname{} outperforms competitive baselines, and finds out explainable causal connections on fMRI datasets.
\end{abstract}

\section{Introduction}
Data science is moving from the data-centric paradigm forward the science-centric paradigm, and causal revolution is sweeping across various research fields. Causality learning endeavors to unearth causal relationships among variables from observational data and generate causal graph, that is, directed acyclic graph (DAG). Unlike correlation-based study, causality analysis reveals the causal mechanism of data generation. Identifying causality holds paramount significance for stable inference and rational decisions in many applications, such as recommendation systems \cite{wang2020recommendation}, medical diagnostics \cite{richens2020medical}, epidemiology \cite{vandenbroucke2016epidemiology} and many others \cite{von2022fairness}.

For its significance, numerous studies have been conducted toward causal structure learning. Constraint-based algorithms \cite{spirtes2000PC,colombo2014PCstable,marella2022PC-cs} acquire a set of causal graphs that satisfy the conditional independence (CI) inherent in the data. Nevertheless, the faithfulness assumption can be refuted, and a substantial number of CI tests are needed. Score-based algorithms \cite{chickering2002GES,hauser2012GIES,ramsey2017FGES} define a structure scoring function in conjunction with search strategies to explore the DAG that best fits the limited data. However, these algorithms cannot differentiate DAGs that belong to the same Markov equivalence class (MEC). By virtue of additional assumptions on data distribution and functional classes, functional causal model-based methods \cite{hoyer2009ANM,zhang2009PNL} can distinguish DAGs within the same MEC, but the exhaustive and heuristic search of DAGs encounters the challenge of combinatorial explosion with the increasing number of variables. NOTEARS \cite{zheng2018NOTEARS} formulates the acyclicity constraint as a differentiable equation and applies efficient numerical solvers to search DAG. Subsequent efforts focus on nonlinear extension \cite{lachapelle2020GraNDAG,zheng2020NOTEARSMLP}, optimization technique \cite{wang2021CORL,yang2023RCLOG},  robustness \cite{he2021DARING} and semi-structured data \cite{liang2023HetDAG}.

However, these algorithms simply deem causal relationships stand exclusively at the level of individual variables (\emph{micro-variable}), ignoring the collective interactions from multiple variables (\emph{macro-variable}). 
For instance, the brain can be characterized at a micro granularity of neurons and their synapses, but high-order synergistic subsystems are widespread, which typically sit between canonical functional networks and may serve an integrative role \cite{varley2023brain}.
Actually, observational data can be regarded as knowledge in the lowest granularity level, while knowledge can be regarded as the abstraction of data at different granularity levels \cite{wang2017DGCC,wang2022GrCreview}. Similar viewpoints appear in the research of complex systems, which suggests that causal relationship is more pronounced at the macro-scale of a system than at its micro-scale. This phenomenon, widely known as \emph{causal emergence}, manifests extensively across scientific domains such as physics \cite{loewer2012physicsemergence}, physiology \cite{noble2012physiologyemergence}, sociology \cite{elder2012sociologyemergence}, and beyond. Due to the intricacy of macro-level causality, available algorithms often overlook or misinterpret the underlying causal structure. Furthermore, the high complexity of causal discovery algorithms causes a significant efficacy drop when dealing with high-dimensional data, hindering their practical implementations.

\begin{figure}[ht!]
\centering
\includegraphics[width=8cm,height=6.5cm]{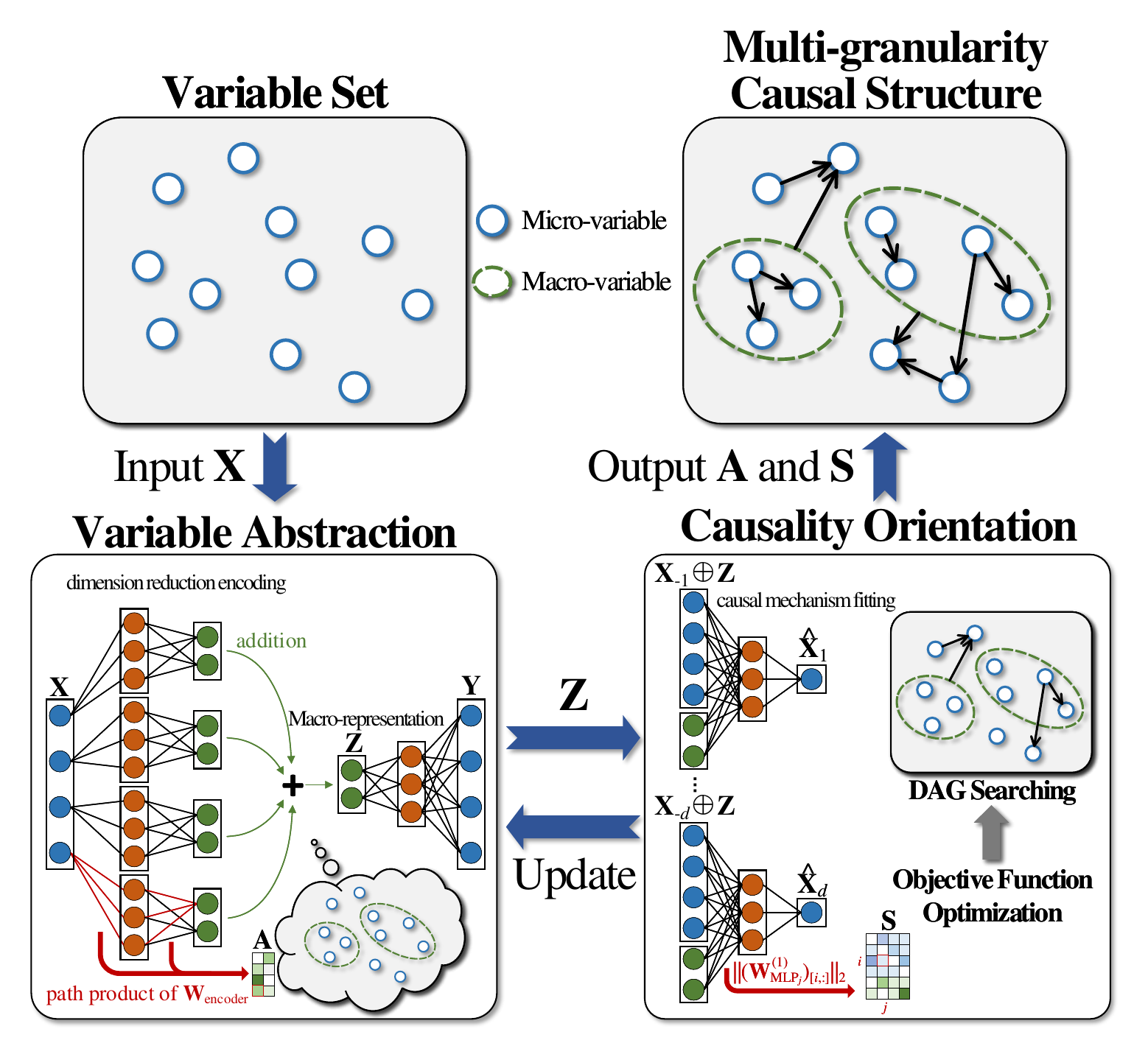}
\caption{Framework overview of \methodname{}, which takes observational data $\mathbf{X}$ as inputs and constructs an SAE to explore coarse-grained strategies and causal abstractions. Within the encoder, each input variable is trained individually, and their encoded representations are summed to obtain the latent macro-variable representation $\mathbf{Z}$, which is then used by the decoder to reconstruct the observational data $\mathbf{X}$. The contribution matrix $\mathbf{A}$ from micro-variables to macro-variables is extracted from the path product of the encoder parameters $\mathbf{W}_\text{encoder}$. Next, \methodname{} feeds the concatenation of micro- and macro-variables $\mathbf{X}\oplus\mathbf{Z}$ into each MLP to explore potential causal relationships, and collects the multi-granularity weighted adjacency matrix $\mathbf{S}$ from the first parameter matrix of MLPs $\mathbf{W}_\text{MLP}^{(1)}$. 
\methodname{} formulates the DAG search as an optimization problem with a simplified acyclic constraint, and induces the multi-granularity causal structure from $\mathbf{S}$ and $\mathbf{A}$.}
\label{fig:framework}
\end{figure}

We aim to learn multi-granularity causal structure and propose an approach called \methodname{}, as depicted in Figure \ref{fig:framework}. \methodname{} firstly establishes a sparse autoencoder (SAE) \cite{ng2011SAE} to automatically coarse-grain the micro-variables into latent macro-ones. Next, it constructs a multiple layer perceptron (MLP) for each micro-variable, taking both the micro- and macro-variables as inputs to explore the underlying causal mechanisms. It further introduces a simplified acyclicity constraint to efficiently orient  edges. 
The major contributions of our work are:\\
\noindent (i) Diverging from existing causal learning methods that exclusively focus on causality at the micro-level of individual variables, we  pioneer to learn multi-granularity causal structure across micro- and macro-variables, which is more complex and challenging but holds immense practical values in various scenarios.\\
\noindent (ii) Our proposed \methodname{} utilizes SAE with sparsity term on the encoder to explore potential macro-variables and extract effective coarse-graining strategies in a sensible way. \methodname{} leverages MLPs to model the underlying causal mechanisms and introduces a simplified acyclicity penalty to orient edges, searching causal structure in an efficacy way.\\
\noindent (iii) Experimental results confirm the advantages of \methodname{} over  competitive baselines \cite{spirtes2000PC,chickering2002GES,yu2019DAGGNN,Ng2019GAE,lachapelle2020GraNDAG,zheng2020NOTEARSMLP}, and it identifies multi-granularity causal connections from fMRI data.

\section{Related Work}
{Causal structure learning is an indispensable and intricate task pervading in various scientific fields. Existing causal learning algorithms can be grouped into two types: constraint-based and score-based. Constraint-based methods  \cite{spirtes2000PC, bird2008hippocampus,marella2022PC-cs} perform CI tests to obtain causal skeleton, then orient edges via elaborate rules to meet the requirements of DAG. Score-based solutions \cite{chickering2002GES, hauser2012GIES, ramsey2017FGES} leverage the score function and  search strategies to find the graph yielding the highest score.  NOTEARS \cite{zheng2018NOTEARS} recasts the combinatoric graph search problem as a continuous optimization problem, stimulating a proliferation of literature. DAG-GNN \cite{yu2019DAGGNN} and GAE \cite{Ng2019GAE} extend NOTEARS to nonlinear cases using autoencoder. GraN-DAG \cite{lachapelle2020GraNDAG} and NOTEARS-MLP \cite{zheng2020NOTEARSMLP} employ MLPs to approximate the underlying data generation functions. DARING \cite{he2021DARING} imposes residual independence constraint in an adversarial way to facilitate DAG learning. RCL-OG \cite{yang2023RCLOG} uses order graph instead of Markov chain Monte Carlo to approximate the posterior distribution of DAGs. However, these causal solutions only focus on the causality among micro-variables, can not cope with multiplex causality at different granularity levels. Our \methodname{} bridges this gap by leveraging SAE to explore multi-granularity causality between micro- and macro-variables, and gradient-based search with simplified acyclicity constraint to seek DAG in an efficacy way.
}

{Actually, the behavior and properties of a system are regulated by causal relations at different granularity levels. The micro-level causality cannot reveal the complexity of a system, especially for the cross-regulations among different levels. Moreover, the extracted macro-variables represent a form of knowledge that unveils higher-level characteristics and collective behavioral patterns within a system. Inspired by the human  cognitive principle   `from coarser to finer', granular cognitive computing aims to process information at various granularity levels  \cite{wang2017DGCC,wang2022GrCreview}. Causal learning from the perspective of granular computing has received much less attention, compared with that from individual micro-variables. {MaCa \cite{yang2022MaCa} models the text with multi-granularity way and uses a matrix capsule network to cluster the emotion-cause pairs. MAGN \cite{qiang2023MAGN} performs backdoor adjustment based on structural causal model to achieve cross-granularity few-shot learning.} EMGCE \cite{wu2023EMGCE} represents sentences at different granulation layers to extract explicit and implicit causal triplets of words or phrases. 
While \methodname{} focuses on exploring the multi-granularity causal graph among micro- and macro-variables. 
}

\section{The Proposed \methodname{} Algorithm}
\subsection{Problem Definition}
In this section, we introduce main concepts used in this paper, followed by a formal definition of our problem.

\begin{definition}[Structural Causal Model (SCM)] 
The SCM is defined on a set of variables $\mathcal{V}$$=$$(v_1,v_2,\cdots,v_d)$, and consists of a causal graph $\mathcal{G}$$=$$(\mathcal{V},\mathcal{E})$ along with structural equations.  Each edge $(i,j)$$\in$$\mathcal{E}$ represents a direct causal relation from $v_i$ to $v_j$. $v_i$ represents a micro-variable, with its observational data denoted as $\mathbf{x}_i$. Multiple micro-variables can be abstracted into a macro-variable $u_i$, with its representative data denoted as $\mathbf{z}_i$. The distribution $P$ is said to be Markov with respect to $\mathcal{G}$, allowing the joint probability $P$ to be decomposed into the product of conditional probabilities as $P(\mathbf{X})$$=$$\prod_{i=1}^dP(\mathbf{x}_i|\mathbf{x}_{pa(i)})$, where $\mathbf{x}_{pa(i)}$ acts as the set of parents of $\mathbf{x}_i$. We assume that the distribution $P$ is entailed by Additive Noise Models (ANMs) \cite{peters2014ANM} of the form:
\begin{equation}
    \mathbf{x}_i=f_i(\mathbf{x}_{pa(i)})+N_i
\end{equation}
where $f_i$ is a nonlinear function that denotes the generative process of $\mathbf{x}_i$, $N_i$ is the external noise of $\mathbf{x}_i$.
\end{definition}

\begin{definition}[Directed acyclic graph (DAG)] 
A DAG consists of variables and edges, with each edge directed from one variable to another. For instance, $v_i\to v_j$ means there is a directed edge from $v_i$ to $v_j$ and hence $v_i$ is a direct cause of $v_j$. A path between $v_i$ and $v_j$ in a DAG is a sequence of edges. The ending variable of each edge acts as the starting variable of the next edge in the sequence. If there is no path from any variable to itself, then the graph is acyclic. In this paper, we further extend the concept of DAG to encompass multi-granularity causal structures, wherein if there exists an edge $u_i\to v_j$, there should be no path from $v_j$ to any micro-variable composing $u_i$.
\end{definition}

\begin{problem}
Given observational data $\mathbf{X}$=$(\mathbf{x}_1,\mathbf{x}_2,$$\cdots$$,\mathbf{x}_d)$, our task is to uncover multi-granularity causal DAG. This DAG can reflect the collective behavior of variable clusters and reveal complex  causal interactions.
\end{problem}

\subsection{Variable Abstraction}
Abstract descriptions provide the foundation for system interventions and explanations for observed phenomena at a coarser granularity level than the most fundamental account of the system. While the emerging causal representation learning aim to reconstruct disentangled causal variable representations from unstructured data (e.g. images) \cite{scholkopf2021representation}, our intention is to learn from observational data and generate low-dimensional representations of macro-variables, which typically encapsulate high-order interactions and collective behaviors among multiple micro-ones.
SAE is a unsupervised learning neural network that excels in extracting compact and meaningful representations of data. By promoting sparsity, it encourages the activation of only a small subset of neurons in the hidden layer, which not only reduces the data dimensionality but also enhances interpretability. 

Given that, \methodname{} includes an SAE-based module to gain causal abstraction from micro-variables to macro-ones. However, conventional SAE uses a shared encoder for all inputs, entangling the contributions of micro-variables to macro-ones. To ensure the independence of micro inputs during the coarse-graining process, we construct $d$ encoders to separately encode each input, and define the encoder and decoder as:
\begin{equation}
\begin{aligned}
\label{SAE}
\mathbf{Z}&=\sum_{i=1}^d\sigma(\sigma(\mathbf{X}_{[:,i]}(\mathbf{W}_\text{encoder}^{(1)})_{[i,:,:]})(\mathbf{W}_\text{encoder}^{(2)})_{[i,:,:]})\\
\mathbf{Y}&=\sigma(\sigma(\mathbf{Z}\mathbf{W}_\text{decoder}^{(1)})\mathbf{W}_\text{decoder}^{(2)})
\end{aligned}
\end{equation}
where $\mathbf{Z}$$=$$(\mathbf{z}_1,\mathbf{z}_2,\cdots,\mathbf{z}_q)$ represents the encoded data of the latent macro-variables, $\sigma$ is an activation function, and biases are omitted for clarity. $\mathbf{W}_\text{encoder}^{(1)}$$\in$$\mathbb{R}^{d\times 1\times m1}$, $\mathbf{W}_\text{encoder}^{(2)}$$\in$$\mathbb{R}^{d\times m1\times q}$, $\mathbf{W}_\text{decoder}^{(1)}$$\in$$\mathbb{R}^{q\times m1}$, $\mathbf{W}_\text{decoder}^{(2)}$$\in$$\mathbb{R}^{m1\times d}$ are parameters of encoder and decoder, $m_1$ is the number of neurons in the first hidden layer. Although we consider all micro-variables as inputs, in our coarse-grained strategies, not every micro-variable contributes to the generation of macro-variable representations. To determine which micro-variables make contribution, we define the contribution as the path product on the encoder:
\begin{equation}
\label{matrixA}
    \mathbf{A}=\vert\mathbf{W}_\text{encoder}^{(1)}\vert\vert\mathbf{W}_\text{encoder}^{(2)}\vert
\end{equation}

We say that a path is inactive if at least one weight along the path is zero. Therefore, when the path product from input $i$$=$$1,2,\cdots,d$ to output $j$$=$$1,2,\cdots,q$ is non-zero, i.e. $\mathbf{A}_{i,j}$$\neq$$0$, $v_i$ is one of the constituents of macro-variable $u_j$. To clarify the composition of macro-variables from micro-ones, we introduce an $l_{1,1}$-norm regularization on the parameter matrices of the encoder, which encourages sparsity of $\mathbf{A}$. This allows some $\mathbf{z}_i$$\in$$\mathbf{Z}$ to approach the zero vector, achieving dynamic variability in the number of macro-variables. In addition, we incorporate a data reconstruction loss, which helps retain causal interpretation. In other words, the compressed macro-variables  preserve a significant portion of essential causal information, allowing for the faithful reconstruction of the original data. 
Then the loss function of variable abstraction module is defined as follows:
\begin{equation}
\label{loss1}
    \mathcal{L}_1=\mathcal{J}(\mathbf{X},\mathbf{Y})+\alpha_1\Vert\mathbf{W}_{\text{encoder}}\Vert_{1,1}
\end{equation}
where $\mathcal{J}(\cdot)$ denotes the cost function, here we use mean squared error, that is $\mathcal{J}(\mathbf{X},\mathbf{Y})$$=$$\frac{1}{2n}\Vert\mathbf{X}-\mathbf{Y}\Vert_F^2$, $n$ is the number of samples. 
The usage of the regularization term may lead to a reduction of variables contributing to abstractions. In such cases, to ensure approximate data reconstruction, SAE prefers preserving the inputs of parent variables, as they encode more information than child ones.
As above, \methodname{} extracts the macro-representation vectors $\mathbf{Z}$ from $\mathbf{X}$, which will be used for subsequent multi-granularity causality orientation.

\subsection{Causality Orientation}
How to orient causal edges between variable nodes is a critical task in causal discovery. Early approaches determine the orientation based on specific structures or asymmetry. The differentiable acyclicity constraint of NOTEARS \cite{zheng2018NOTEARS} enables the usage of standard numerical techniques to search DAGs. However, NOTEARS and its variants are typically operated at the micro-level. In this work, we need to identify nonlinear dependencies among variables and enforce acyclicity across multiple granularities.  For this purpose, we propose a simplified acyclicity constraint to improve search efficiency.
We first construct an MLP for each micro-variable, taking other variables and latent macro-variables as inputs, to model the underlying causal mechanisms:
\begin{equation}
\label{MLP}
    \hat{\mathbf{x}}_i=\text{MLP}_i(\mathbf{X}_{-i}\oplus\mathbf{Z};\mathbf{W}_{\text{MLP}_i})
\end{equation}
where $\oplus$ is the concatenation operator and $\mathbf{W}_{\text{MLP}_i}$ is parameters of the $i$-th MLP. Actually, the variables involved in the fitting process of $\mathbf{x}_i$ are potential parent variables of $v_i$. Therefore, we extract the weighted adjacency matrix (WAM) $\mathbf{C}$ on the first layer of $\mathbf{W}_{\text{MLP}_i}$, taking into account the macro-variables:
\begin{equation}
\label{matrixC}
    \mathbf{C}_{i,j}=\Vert(\mathbf{W}_{\text{MLP}_j}^{(1)})_{[i,:]}+\mathbf{A}_{[i,:]}(\mathbf{W}_{\text{MLP}_j}^{(1)})_{[d+1:,:]}\Vert_2
\end{equation}
where $\mathbf{W}_{\text{MLP}_j}^{(1)}$$\in$$\mathbb{R}^{(d+q)\times m_2}$ represents the first parameter matrix of $j$-th MLP, $m_2$ is the number of neurons in the first hidden layer. Consequently, we can collectively enforce acyclicity on the multi-granularity variables. Similarly, we can obtain multi-granularity WAM $\mathbf{S}$$\in$$\mathbb{R}^{(d+q)\times d}$ as:
\begin{equation}
\label{matrixS}
    \mathbf{S}_{i,j}=\Vert(\mathbf{W}_{\text{MLP}_j}^{(1)})_{[i,:]}\Vert_2
\end{equation}

When a variable participates in joint interactions on other variables, we reduce its individual weight on other variables to exclude redundant causal relations by a redundancy penalty as:
\begin{equation}
\label{penalty}
\begin{aligned}
    \mathcal{L}_{red}=\sum_{j=1}^d\Vert&\sum_{k=1}^{m_2}(\vert\mathbf{A}\vert\vert\mathbf{W}_{\text{MLP}_j}^{(1)}\vert_{[d+1:,:]})_{[:,k]}\circ\\
    &\sum_{k=1}^{m_2}(\vert\mathbf{W}_{\text{MLP}_j}^{(1)}\vert_{[:d,:]})_{[:,k]}\Vert_1
\end{aligned}
\end{equation}
where $\circ$ denotes the Hadamard product. 

Then we can define the loss function of causality orientation module as follows:
\begin{equation}
\begin{aligned}
        \min\quad&\mathcal{L}_2=\mathcal{J}(\mathbf{X},\hat{\mathbf{X}})+\mathcal{L}_{red}+\\
        &\alpha_2(\sum_{j=1}^d\Vert\mathbf{W}_{\text{MLP}_j}^{(1)}\Vert_{1,1}+\frac{1}{2}\sum_{j=1}^d\Vert\mathbf{W}_{\text{MLP}_j}\Vert_F^2)\\
        \text{s.t.}\quad&\mathcal{H}(\mathbf{C})=0
\end{aligned}
\end{equation}
${H}(\mathbf{C})=0$ is our meticulously simplified  directed acyclicity constraint, which will be discussed in more detail later.
As sparse parameters lead to more specific direction, we apply an $l_{1,1}$-norm regularization on each $\mathbf{W}_{\text{MLP}_j}^{(1)}$ to help orientation. We further employs squared Frobenius norm on each $\mathbf{W}_{\text{MLP}_j}$ to facilitate model convergence and enhance generalization. The objective function of \methodname{} consists of the two afore-defined loss functions:
\begin{equation}
\begin{aligned}
        \label{obj}
        \min\quad&\mathcal{L}=\mathcal{L}_1+\mathcal{L}_2\\
        \text{s.t.}\quad&\mathcal{H}(\mathbf{C})=0
\end{aligned}
\end{equation}
By iteratively optimizing the objective function, we can search for the optimal multi-granularity DAG. Due to the continuous nature of $\mathbf{C}$ and $\mathbf{S}$ obtained after training, it cannot be directly interpreted as a causal graph. Therefore, post-processing procedure on $\mathbf{C}$ and $\mathbf{S}$ is required. We first cut all weights below a certain threshold $\epsilon$, then we iterative eliminate the smallest weight until no directed cycles exist in $\mathbf{C}$ and $\mathbf{S}$. Finally, we set all remaining non-zero elements to $1$.
\subsection{Optimization}

To ensure the acyclicity of $\mathbf{C}$, an explicit way is to enforce the following condition: $\sum_{k=1}^d\text{tr}(\mathbf{D}^k)$$=$$0$ where $\mathbf{D}$$=$$\mathbf{C}$$\circ$$\mathbf{C}$ and $\text{tr}(\cdot)$ denotes the trace of $\mathbf{D}$. NOTEARS \cite{zheng2018NOTEARS} suggests that $\sum_{k=1}^d\text{tr}(\mathbf{D}^k)$$=$$0\Rightarrow\text{tr}(\sum_{k=1}^d\frac{\mathbf{D}^k}{k!})$$=$$0\Rightarrow\text{tr}(\sum_{k=0}^d\frac{\mathbf{D}^k}{k!})-d$$=$$0$, and proposes Theorem \ref{theorem:notears}.

\begin{theorem}
\label{theorem:notears}
A WAM $\mathbf{C}\in\mathbb{R}^{d\times d}$ is a DAG if and only if
\begin{equation}
    h(\mathbf{C})=\text{tr}(e^{\mathbf{D}})-d=0
    \label{notears}
\end{equation}
\end{theorem}
This provides a continuous optimization approach to search for the graph structure. Indeed, the computational complexity of matrix exponential makes it challenging to obtain results within an acceptable time on high-dimensional data \cite{liang2023GranLCS}. Here, we introduce an meticulously designed acyclicity constraint to orient edges in an efficacy way.

\begin{proposition}
\label{eigval}
A WAM $\mathbf{C}\in\mathbb{R}^{d\times d}$ is a DAG if and only if 
\begin{equation}
   \forall i\in[1,d], \lambda_i=0
\end{equation}
where $\lambda_i$ represents the eigenvalues of $\mathbf{D}$.
\end{proposition}
The proof of Proposition \ref{eigval} is deferred into the Supplementary file. {Previous effort \cite{lee2019NOBEARS} attempts to enforce acyclicity based on the spectral radius, which is unstable and tends to a zero matrix, due to its neglect of global information of eigenvalues and an easily attainable local optimum of the zero matrix.} 
Optimizing the eigenvalues jointly expands the search space but introduces computational burdens. Since a square matrix is necessarily unitarily similar to an upper triangular matrix, we perform \emph{Schur decomposition} on $\mathbf{D}$:
\begin{equation}
    \mathbf{D}=\mathbf{U}\mathbf{R}\mathbf{U}^\top
\end{equation}
where $\mathbf{U}$ is a unitary matrix, $\mathbf{R}$ is a upper triangular matrix that has the eigenvalues on the diagonal, which are also the eigenvalues of $\mathbf{D}$, since $\mathbf{D}$ is similar to $\mathbf{R}$. 
\begin{proposition}
A WAM $\mathbf{C}\in\mathbb{R}^{d\times d}$ is a DAG if and only if 
\begin{equation}
\label{HC}
    \mathcal{H}(\mathbf{C})=\Vert \text{diag}(\mathbf{R})\Vert_2^2=0
\end{equation}
where $\text{diag}(\cdot)$ denotes the diagonal vector of a matrix.
\end{proposition}

{Optimizing Eq. (\ref{HC}) avoids the intricate computation of matrix exponential and its gradient in Eq. (\ref{notears}), which facilitate an efficient search of causal graphs in the latent graph space. It can also minimize the weights of non-existing edges with the help of reconstruction loss on $\mathbf{X}$.} Therefore, the error caused by transforming $\mathbf{D}$ to $\mathbf{R}$ can be effectively removed through our post-processing procedure.

To this end, Eq. \eqref{obj} is reformulated into a continuous optimization problem, albeit non-convex due to the non-convex feasible set. Nevertheless, we can employ an augmented Lagrangian approach to replace the original constrained problem Eq. (\ref{obj}) with a sequence of unconstrained subproblems, and then seek stationary points as the solution:
\begin{equation}
\begin{aligned}
        \min\quad&\mathcal{L}=\mathcal{J}(\mathbf{X},\mathbf{Y})+\alpha_1\Vert\mathbf{W}_{\text{encoder}}\Vert_{1,1}+\\
        &\mathcal{J}(\mathbf{X},\hat{\mathbf{X}})+\mathcal{L}_{red}+\alpha_2(\sum_{j=1}^d\Vert\mathbf{W}_{\text{MLP}_j}^{(1)}\Vert_{1,1}+\\
        &\frac{1}{2}\sum_{j=1}^d\Vert\mathbf{W}_{\text{MLP}_j}\Vert_F^2)+\frac{\mu}{2}\vert\mathcal{H}(\mathbf{C})\vert^2+\gamma\mathcal{H}(\mathbf{C})
\end{aligned}
\end{equation}
where $\mu$ is the penalty coefficient and $\gamma$ is the Lagrange multiplier. The parameter update process is as follows:
\begin{align}
\label{updparam}
\nonumber&\boldsymbol{\theta}^{(\kappa+1)}\gets\arg\min\{\mathcal{J}(\mathbf{X},\mathbf{Y})+\alpha_1\Vert\mathbf{W}_{\text{encoder}}\Vert_{1,1}+\\
\nonumber&\mathcal{J}(\mathbf{X},\hat{\mathbf{X}})+\mathcal{L}_{red}+\alpha_2(\sum_{j=1}^d\Vert\mathbf{W}_{\text{MLP}_j}^{(1)(\kappa)}\Vert_{1,1}+\\
        &\frac{1}{2}\sum_{j=1}^d\Vert\mathbf{W}_{\text{MLP}_j}^{(\kappa)}\Vert_F^2)+\frac{\mu}{2}\vert\mathcal{H}(\mathbf{C})\vert^2+\gamma\mathcal{H}(\mathbf{C})\}\\
\label{updmu}
&\mu^{(\kappa+1)}\gets
\left\{
\begin{aligned}
&\eta\mu^{(\kappa)}, && \mathcal{H}(\mathbf{C}^{(\kappa+1)})>\rho\mathcal{H}(\mathbf{C}^{(\kappa)})\\
&\mu^{(\kappa)},     && \text{otherwise}
\end{aligned}
\right.\\
\label{updgamma}
&\gamma^{(\kappa+1)}\gets\gamma^{(\kappa)}+\mu\mathcal{H}(\mathbf{C})
\end{align}
where $\boldsymbol{\theta}\triangleq\{\mathbf{W}_\text{encoder},\mathbf{W}_\text{decoder},\mathbf{W}_\text{MLP}\}$, and $\boldsymbol{\theta}^{(\kappa)}$ is the iterative optimized $\boldsymbol{\theta}$ in the $\kappa$-th iteration. 
A variety of optimization methods can be applied to Eq. (\ref{updparam}). In this work, we apply the L-BFGS-B algorithm \cite{byrd1995lbfgsb}. The comprehensive procedure of \methodname{} is delineated in Algorithm 1 in the Supplementary file.

The time complexity of \methodname{} comes from two sources, variable abstraction and causality orientation. For the former, the computational complexity of SAE for one iteration step is $\mathcal{O}(nm_1q+ndm_1q+nd^2m_1)$. For the latter, the computational complexity of MLP for one iteration step is $\mathcal{O}(d^2m_2+nd^2m_2+d^3)$. {Although the \emph{Schur decomposition} of \methodname{} and the trace of matrix exponential in NOTEARS share the same time complexity of $\mathcal{O}(d^3)$, eigenvalues are more sensitive to the presence of directed cycles in $\mathbf{C}$ than the trace of matrix exponential, enabling $\mathbf{C}$ to be optimized faster towards a cycle-free direction. This is because under the constraint of the squared Frobenius norm, most entries of $\mathbf{C}$ are smaller than 1, resulting in the trace of matrix exponential approaching zero. Therefore, our approach is approximately one order of magnitude faster, as shown in the experiment section.} 

\section{Experimental Evaluation}

\subsection{Experimental Setup}

We evaluated the proposed \methodname{} on random DAG produced by \emph{Erd\H os-R\'enyi} (ER) or \emph{Scale-Free} (SF) scheme. We varied the number of variables ($d$$\in$$\{20, 50, 100\}$) with edge density (degree=2). For each graph, we generate 10 datasets of $n$=1000 samples following: (i) Additive  models with  Gaussian processes, and (ii) ANM with Gaussian processes. In addition, on the SF graph with $d$$=$$20$, we randomly select $\{2,4\}$ variables as macro-variables and employ MLP to decompose them into $8$ micro-variables, forming multi-granularity graphs. We consider another well-known dataset Sachs \cite{sachs2005biology}, which measures the level of different expressions of proteins and phospholipids in human cells. The ground truth causal graph of this dataset consists of 11 variables and 20 edges. In this work, we test on observational data with 7466 samples.

We compare \methodname{} with representative causal structure learning methods, including PC \cite{spirtes2000PC}, GES \cite{chickering2002GES}, DAG-GNN \cite{yu2019DAGGNN}, GAE \cite{Ng2019GAE}, GraN-DAG \cite{lachapelle2020GraNDAG} and NOTEARS-MLP \cite{zheng2020NOTEARSMLP}. The first two methods focus on combinatorial optimization, and the last four target at general nonlinear dependencies.
We refer readers to the Supplementary file for the detailed experimental setup.

We evaluate the estimated DAG using five metrics: Precision: proportion of correctly estimated edges to the total estimated edges; Recall: proportion of correctly estimated edges to the total edges in true graph; F1 score: the harmonic average of precision and recall; Structural hamming distance (SHD): the number of missing, falsely estimated or reversed edges. Runtime: the running time required to obtain the results, measured in seconds. When a method fails to return any result within a reasonable response time (100 hours for a single training in our case), we mark the entry as '-'.
For the reported results, ↑(↓) means the higher (lower) the value, the better the performance is. The best result is highlighted in \textbf{bold} font. 

\subsection{Result Analysis}

\begin{figure}[!b]
\centering
\subfigure[$d$=20 with 2 macro-variables]{\includegraphics[width=8cm]{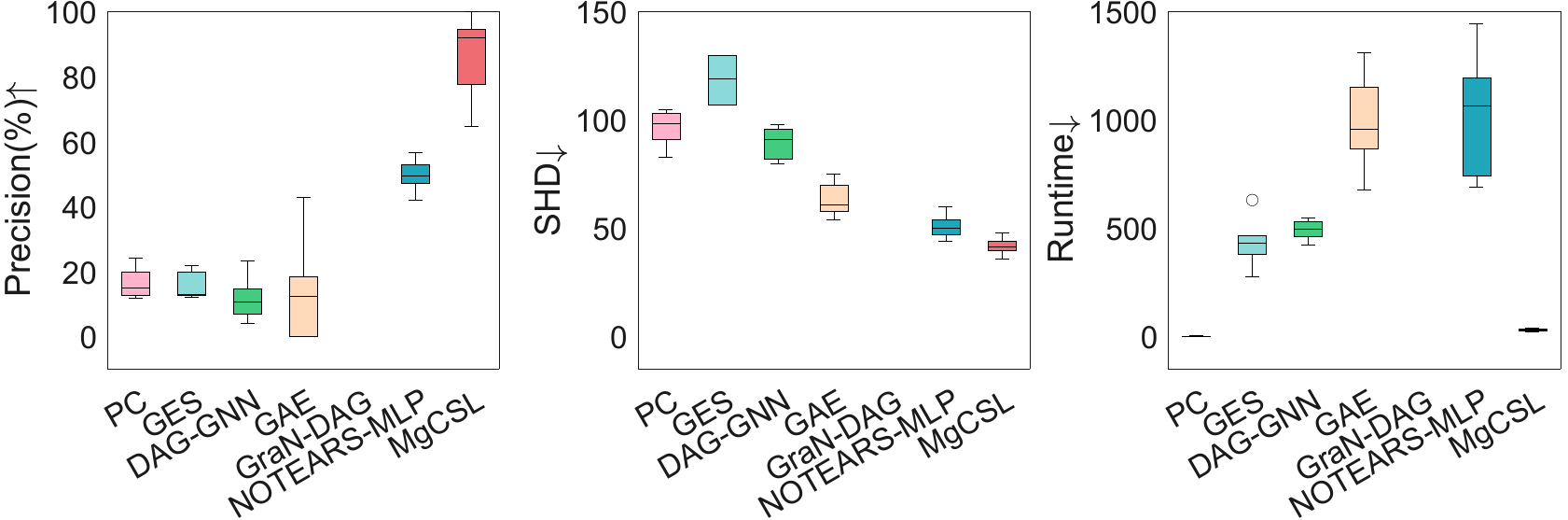}}
\subfigure[$d$=20 with 4 macro-variables]{\includegraphics[width=8cm]{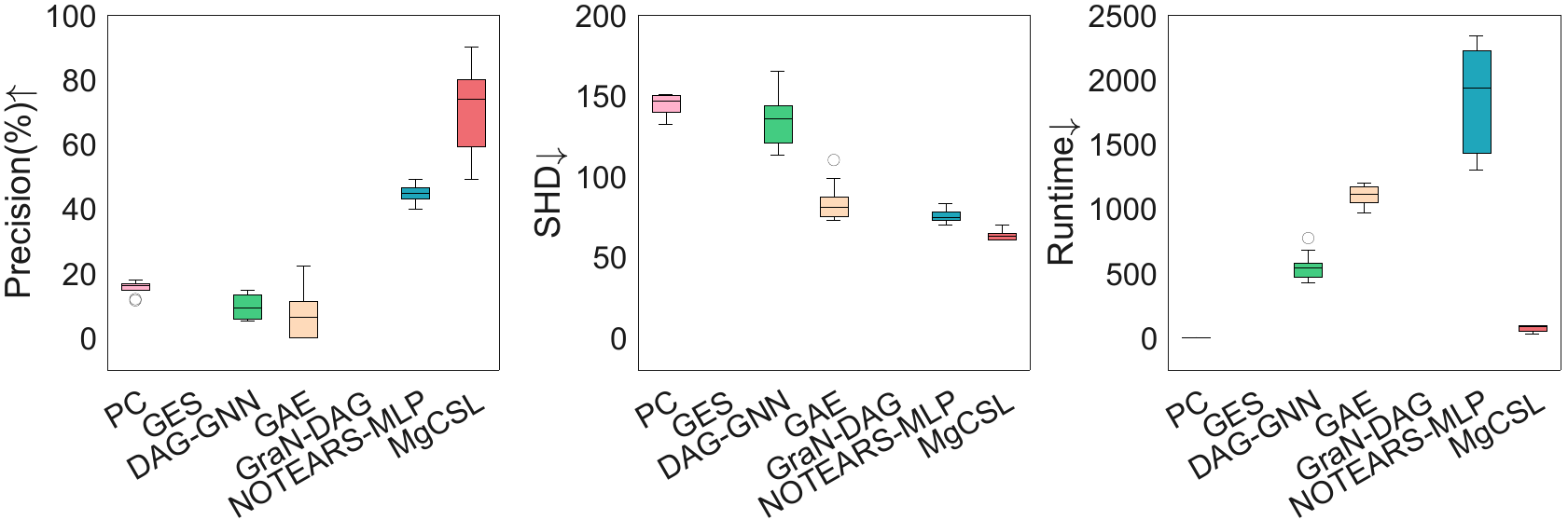}}
\caption{Results on multi-granularity graph.}
\label{fig:Mg}
\end{figure}

\begin{table*}[htbp]
\tiny
\setlength\tabcolsep{3pt} 
\centering
\begin{tabular}{c l| r r r r| r|| r r r r| r}
\hline
      & &\multicolumn{5}{c|}{\emph{Erd\H os-R\'enyi} graph} &\multicolumn{5}{c}{\emph{Scale-Free} graph}\\
\hline
$\#$Variables && Precision(\%)$\uparrow$ & Recall(\%)$\uparrow$ & F1(\%)$\uparrow$ & SHD$\downarrow$ & Runtime(s)$\downarrow$ &  Precision(\%)$\uparrow$ & Recall(\%)$\uparrow$ & F1(\%)$\uparrow$ & SHD$\downarrow$ & Runtime(s)$\downarrow$ \\
\hline
\multicolumn{1}{c|}{\multirow{7}{*}{20}}    
      & PC & 60.22$\pm$8.38 & 54.50$\pm$7.43 & 57.15$\pm$7.57 & 25.90$\pm$4.15 & 0.54$\pm$0.20 & 41.20$\pm$9.26 & 42.70$\pm$9.60 & 41.85$\pm$9.10 & 33.70$\pm$5.23 & 0.69$\pm$0.29  \\
\multicolumn{1}{c|}{} & GES & 58.14$\pm$5.80 & 59.00$\pm$6.15 & 58.48$\pm$5.43 & 26.00$\pm$3.06 & 12.28$\pm$2.01 & 43.72$\pm$5.88 & 55.68$\pm$6.39 & 48.93$\pm$5.87 & 33.60$\pm$3.37 & 14.75$\pm$2.10  \\
      \cline{2-12}
\multicolumn{1}{c|}{} & DAG-GNN & 89.57$\pm$8.44 & 42.50$\pm$6.01 & 57.50$\pm$6.85 & 24.70$\pm$3.09 & 2512.59$\pm$674.56 & 85.02$\pm$11.23 & 38.92$\pm$5.29 & 53.08$\pm$6.06 & 23.90$\pm$2.64 & 2880.47$\pm$358.99  \\
\multicolumn{1}{c|}{} & GAE & 83.24$\pm$12.07 & 30.50$\pm$17.59 & 41.01$\pm$17.84 & 29.30$\pm$5.42 & 1469.75$\pm$210.24 & 83.93$\pm$22.12 & 37.03$\pm$19.49 & 48.76$\pm$22.28 & 24.60$\pm$7.97 & 1999.70$\pm$155.48  \\
\multicolumn{1}{c|}{} & GraN-DAG    & 97.13$\pm$5.20 & 51.25$\pm$6.80 & 66.77$\pm$5.58 & 20.10$\pm$2.85 & 311.23$\pm$2.47 & 95.12$\pm$3.69 & 59.19$\pm$12.32 & 72.32$\pm$10.31 & 16.10$\pm$4.46 & 355.68$\pm$2.62\\
\multicolumn{1}{c|}{} & NOTEARS-MLP & 89.86$\pm$4.59 & \bf 69.75$\pm$7.12 & 78.34$\pm$5.17 & 14.80$\pm$3.05 & 45.35$\pm$11.05  & 80.62$\pm$5.62 & \bf 73.24$\pm$5.32 & 76.66$\pm$4.65 & 15.30$\pm$2.87 & 47.10$\pm$10.39  \\
\multicolumn{1}{c|}{} & \methodname{} & \bf 98.61$\pm$2.30 & 65.75$\pm$8.42 & \bf 78.60$\pm$5.73 & \bf 14.10$\pm$3.21 & 9.57$\pm$1.45  & \bf 96.19$\pm$7.36 & 70.54$\pm$7.48 & \bf 81.08$\pm$5.68 & \bf 12.00$\pm$3.59 & 7.58$\pm$3.31  \\
\hline
\multicolumn{1}{c|}{\multirow{7}{*}{50}}   
      & PC & 37.03$\pm$3.48 & 44.10$\pm$4.70 & 40.25$\pm$3.99 & 103.00$\pm$7.06 & 3.40$\pm$0.39 & 36.05$\pm$3.65 & 41.55$\pm$2.75 & 38.58$\pm$3.18 & 100.20$\pm$7.51 & 4.92$\pm$2.26  \\
\multicolumn{1}{c|}{} & GES & 43.38$\pm$5.43 & 56.30$\pm$6.38 & 48.96$\pm$5.69 & 97.10$\pm$12.01 & 666.43$\pm$187.16 & 40.93$\pm$7.52 & 54.02$\pm$9.21 & 46.53$\pm$8.17 & 98.70$\pm$14.18 & 574.28$\pm$100.81  \\
      \cline{2-12}
\multicolumn{1}{c|}{} & DAG-GNN & 83.62$\pm$5.84 & 30.30$\pm$7.07 & 43.88$\pm$6.89 & 72.10$\pm$5.36 & 1598.46$\pm$148.30 & 79.02$\pm$6.15 & 35.77$\pm$2.92 & 49.15$\pm$3.30 & 66.70$\pm$3.62 & 3361.67$\pm$244.01  \\
\multicolumn{1}{c|}{} & GAE & 84.18$\pm$10.22 & 25.20$\pm$13.66 & 37.12$\pm$16.96 & 77.50$\pm$12.37 & 9542.19$\pm$293.50 & 77.40$\pm$18.77 & 24.43$\pm$11.15 & 36.07$\pm$14.29 & 77.90$\pm$10.61 & 2005.66$\pm$151.96  \\
\multicolumn{1}{c|}{} & GraN-DAG    & 57.95$\pm$20.39 & 42.00$\pm$9.24 & 46.22$\pm$7.58 & 99.00$\pm$33.01 & 1039.47$\pm$169.09 & 54.86$\pm$20.04 & 34.95$\pm$6.56 & 40.34$\pm$5.75 & 101.00$\pm$30.13 & 651.25$\pm$39.00\\
\multicolumn{1}{c|}{} & NOTEARS-MLP & 80.38$\pm$4.96 & \bf 64.80$\pm$4.73 & 71.65$\pm$3.96 & 46.60$\pm$5.21 & 365.86$\pm$47.55  & 76.36$\pm$6.11 & 64.53$\pm$4.89 & 69.83$\pm$4.45 & 48.70$\pm$5.44 & 300.23$\pm$48.18  \\
\multicolumn{1}{c|}{} & \methodname{} & \bf 99.08$\pm$1.64 & 59.80$\pm$4.37 & \bf 74.48$\pm$3.30 & \bf 40.80$\pm$4.02 & 81.52$\pm$22.73 & \bf 95.55$\pm$4.83 & \bf 67.32$\pm$5.48 & \bf 78.76$\pm$3.12 & \bf 34.90$\pm$5.24 & 69.48$\pm$17.56  \\
\hline
\multicolumn{1}{c|}{\multirow{7}{*}{100}}   
      & PC & 31.06$\pm$2.79 & 40.60$\pm$3.16 & 35.19$\pm$2.94 & 235.30$\pm$13.67 & 12.74$\pm$3.75 & 27.39$\pm$2.23 & 38.12$\pm$2.61 & 31.86$\pm$2.32 & 263.00$\pm$13.43 & 27.15$\pm$8.60  \\
\multicolumn{1}{c|}{} & GES & 42.10$\pm$3.77 & 60.78$\pm$5.01 & 49.72$\pm$4.16 & 213.44$\pm$18.53 & 79857.07$\pm$101697.32 & - & - & - & - & -\\
      \cline{2-12}
\multicolumn{1}{c|}{} & DAG-GNN & 88.61$\pm$4.75 & 34.95$\pm$6.12 & 49.72$\pm$5.92 & 135.50$\pm$10.05 & 1637.52$\pm$179.76 & 88.80$\pm$4.10 & 31.22$\pm$5.38 & 45.83$\pm$5.72 & 140.10$\pm$8.17 & 3654.05$\pm$214.81  \\
\multicolumn{1}{c|}{} & GAE & 50.15$\pm$42.26 & 13.80$\pm$10.42 & 21.08$\pm$16.05 & 211.20$\pm$55.11 & 5527.23$\pm$2839.86 & 64.56$\pm$35.86 & 15.94$\pm$8.78 & 24.00$\pm$13.26 & 203.00$\pm$62.33 & 1991.81$\pm$143.13  \\
\multicolumn{1}{c|}{} & GraN-DAG    & 11.70$\pm$3.62 & 33.00$\pm$9.84 & 16.89$\pm$4.64 & 653.50$\pm$208.96 & 2089.45$\pm$164.54 & 6.31$\pm$1.10 & 23.60$\pm$5.11 & 9.90$\pm$1.72 & 828.40$\pm$122.29 & 1575.67$\pm$216.80\\
\multicolumn{1}{c|}{} & NOTEARS-MLP & 80.77$\pm$4.88 & \bf 64.50$\pm$5.13 & \bf 71.58$\pm$3.97 & \bf 95.80$\pm$12.69 & 1923.61$\pm$174.89  & 69.28$\pm$4.38 & \bf 62.44$\pm$3.40 & 65.57$\pm$2.70 & 122.60$\pm$12.27 & 1688.14$\pm$387.37  \\
\multicolumn{1}{c|}{} & \methodname{} & \bf 94.61$\pm$3.33 & 49.65$\pm$7.18 & 64.74$\pm$4.84 & 106.10$\pm$11.53 & 430.81$\pm$136.22  & \bf 89.20$\pm$4.97 & 55.28$\pm$7.36 & \bf 67.77$\pm$4.83 & \bf 98.20$\pm$8.09 & 394.56$\pm$51.28\\
\hline
\end{tabular}
\caption{Results on \emph{nonlinear models with additive Gaussian processes}. }
\label{syntaddgp}
\end{table*}

\subsubsection{Results on multi-granularity graph}
We first compare the performance of our \methodname{} with baselines on multi-granularity synthetic datasets. As shown in Figure \ref{fig:Mg}, the presence of macro-variables has an impact on the precision of the baselines. Even on small graphs, they exhibit an precision as low as 0.5 or even lower. PC and GES achieve the highest SHD, because they struggle to capture complex nonlinear relationships through either CI tests or score functions. GES even fails to produce any results when the number of macro-variables is 4. DAG-GNN, GAE and GraN-DAG possess the capability to learn nonlinear dependencies, but they fail to achieve satisfactory results. In particular, GraN-DAG cannot obtain valid estimated graphs across all datasets. This limitation might be attributed to the introduction of noise through the micro-level representations of macro-variables, which affects the DAG search process. NOTEARS-MLP achieves promising results in terms of SHD. However, its low precision and high execution time still make it challenging to directly apply in multi-granularity graphs. \methodname{} is capable of extracting valuable information from data mixed with multi-granularity variables for DAG learning, thereby achieving the optimal performance in terms of precision and SHD with a modest expenditure of time. We apply the Wilcoxon signed-rank test \cite{demvsar2006wilcoxonsignedrank} to check the statistical difference between \methodname{} and other compared methods across these metrics and datasets, all the $p$-value are smaller than 0.02. This demonstrates its ability to discover multi-granularity causal structures.

\subsubsection{Results on typical causal graph}
We also conduct experiments on typical causal discovery synthetic datasets. As shown in Table \ref{syntaddgp} and S2 in the Supplementary file, our \methodname{} outperforms its opponents across most metrics within a short period of time.

\textbf{\methodname{} vs. combinatorial optimization methods:}
PC can swiftly produce comparable results through CI tests with a few variables. However, as the number of variables increases, the limited sample size renders CI tests unreliable, leading to a sharp rise of SHD. The performance of GES is relatively superior to that of PC because it is less affected by data issues (e.g., noisy or small samples). However, its adoption of the greedy search strategy makes it challenging to deliver results within a reasonable runtime, particularly when confronted with the exponential growth of the search space as the number of variables increases. In comparison, our \methodname{} manifests more remarkable performance across these metrics on datasets of different sizes.

\textbf{\methodname{} vs. gradient-based methods:}
DAG-GNN achieves consistent results in most cases by leveraging the power of variational inference. However, it encounters a substantial time cost on low-dimensional data, which hinders its applicability to small graphs. GAE presents poor performance, with its precision fluctuating significantly. While GraN-DAG exhibits excellent precision on small graphs, its performance falters as the number of variables increases. Its estimated graphs contain numerous erroneous edges, indicating its limitation in learning causal structures on large graphs. NOTEARS-MLP shows commendable performance, but it demands a lot of time cost due to the high complexity involved in matrix exponential calculations. \methodname{} surpasses the above rivals across most metrics, maintaining a high precision even on high-dimensional graphs. Moreover, the simplified acyclicity penalty empowers \methodname{} to provide effective results within shorter time, ensuring its feasibility for real-world applications.

\textbf{The impact of different graphs:}
We further compared the performance of the algorithms on ER and SF graphs. On ER graphs, the degrees of individual variables are relatively balanced, whereas SF graphs contain a few hub variables with degrees significantly higher than the average, making it more likely for multiple variables to jointly influence a single target. As the experimental results show, most algorithms such as PC, GES, and NOTEARS-MLP perform worse on SF graphs than that on ER graphs in terms of SHD, especially as the number of variables increases. This could be attributed to the causal discovery algorithm's demand for sparsity in the estimated graph, resulting in poor performance when dealing with graphs with highly uneven degrees. \methodname{} has the capability of abstracting multiple micro-variables into a macro-one and discovering causal edges among multi-granularity variables. This empowers it to achieve competitive results on SF graphs.

\subsubsection{Results on real-world data}

To further verify the effectiveness of \methodname{}, we conduct experiments on Sachs's \cite{sachs2005biology} protein signaling dataset. The results in Table \ref{tab:sachs} show that \methodname{} still exhibits to a competitive performance compared with the rivals. PC, GES and DAG-GNN identify quit a lot of correct edges but at the cost of precision, resulting in the highest SHD. GAE and NOTEARS-MLP miss too many edges, leading to a low recall. The sparse inference graph of GraN-DAG contains fewer errors. This can be attributed to its preprocessing procedure that selects candidate neighbors for each variable, reducing the possibility of making mistakes. \methodname{} outperforms the baselines in terms of precision, F1 and SHD while identifying 6 correct causal edges. This signifies its tangible potential for real-world applications. 

\begin{table}[htbp]
\scriptsize
\centering
\begin{tabular}{l| r r r r}
\hline
            & Precision(\%)$\uparrow$ & Recall(\%)$\uparrow$ & F1(\%)$\uparrow$ & SHD$\downarrow$ \\
\hline
PC & 37.04 & \bf 50.00 & 42.55 & 24\\
GES & 23.08 & 45.00 & 30.51 & 29\\
DAG-GNN & 26.92 & 35.00  & 30.43 & 29 \\
GAE & 28.57 & 10.00 & 14.81 & 20 \\
GraN-DAG  & 50.00 & 10.00 & 16.67 & 18 \\
NOTEARS-MLP & 16.67 & 10.00 & 12.50 &  22\\
\methodname{} & \bf 100.00 & 30.00 & \bf 46.15 & \bf 14\\
\hline
\end{tabular}
\caption{Results on Sachs's protein signaling dataset.}
\label{tab:sachs}
\end{table}

\subsection{Ablation Study}

\begin{figure}[h!tbp]
\centering
\includegraphics[width=8cm]{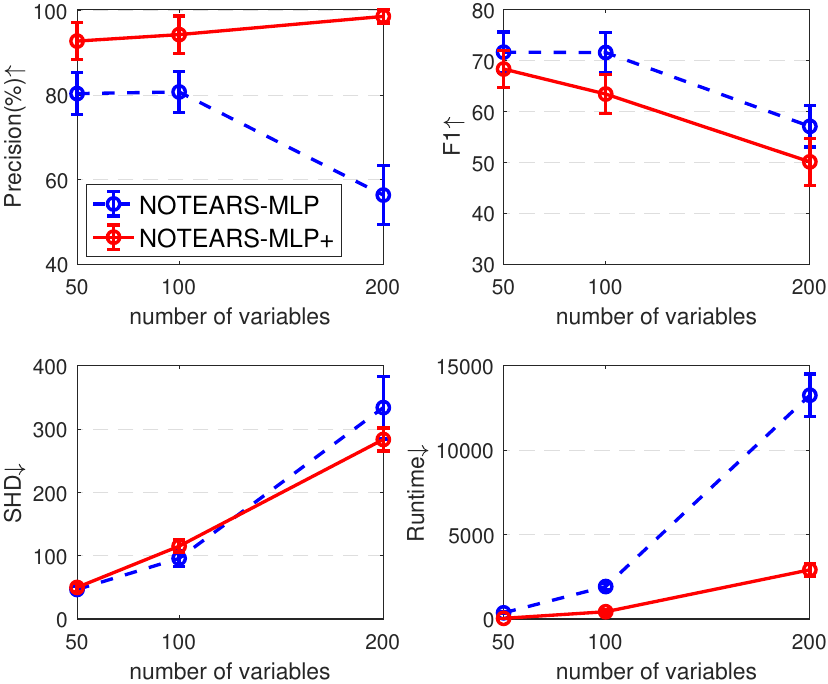}
\caption{Precision, F1, SHD and runtime of NOTEARS-MLP and NOTEARS-MLP combined with our proposed acyclicity constraint named NOTEARS-MLP+ on $d$$=$$\{50,100,200\}$.}
\label{fig:acyclic}
\end{figure}

We conduct ablation study to investigate the effectiveness of the proposed acyclicity constraint. We implement NOTEARS-MLP and NOTEARS-MLP combined with our proposed acyclicity constraint named NOTEARS-MLP+ on the dataset of nonlinear models with additive Gaussian processes. As shown in Fig. \ref{fig:acyclic}, NOTEARS-MLP+ yields sustained high precision even as the number of variables increases, while NOTEARS-MLP discovers excessive erroneous edges. Due to the stronger influence of directed cycles on eigenvalues, NOTEARS-MLP+ prunes a significant number of edges, including correct ones, leading to a decrease in F1 score. Nevertheless, in terms of SHD, NOTEARS-MLP+ manifests comparable performance to NOTEARS-MLP. Most importantly, NOTEARS-MLP+ demonstrates a significant decrease in time consumption, about an order of magnitude. This verifies that the proposed acyclicity penalty is beneficial in efficiently and accurately discovering causal DAG.

In addition, we carry out parameter sensitivity study w.r.t. $\alpha_1$ and $\alpha_2$. A small $\alpha_1$ may introduce considerable noise, while an excessively large $\alpha_1$ could weaken the representational capacity of macro-variables. Besides, a larger $\alpha_2$ can improve the performance in terms of precision and running time, but \methodname{} tends to achieve an overly sparse estimated graph with a few edges when $\alpha_2$ is too large. Based on the aforementioned analysis, we set $\alpha_1$=0.1 and $\alpha_2$=0.01.

\subsection{Case Study}

To further appraise the effectiveness of \methodname{} in real-world applications, we conducted a case study on the functional magnetic resonance imaging (fMRI) Hippocampus dataset, which is consisted of the resting-state signals from six separate brain regions  \cite{poldrack2015fMRI}. It was collected from the same person for 84 days and the sample size for each day is 518. We apply our proposed \methodname{} on 10 successive days. We consider the top 5 edges discovered by \methodname{} as the estimated causal edges: \{PHC, Sub\}$\rightarrow$ERC, ERC$\rightarrow$DG, CA1$\rightarrow$Sub, and CA1$\rightarrow$DG. We compared the results with anatomical directed connections \cite{bird2008hippocampus,zhang2017CDNOD}, since theoretically, when two regions have an anatomical connection, there is a high likelihood of causal relationship between them. The ground truths provided by anatomical structures contain cycles, while we assume that there are no directed cycles in the causal graph. The results  indicate that the first four edges discovered by \methodname{} match the fact that most of the hippocampus's neocortical inputs come from the PHC via the ERC, with most of its neocortical output directed towards the Sub, which also projects back to the ERC. The collaborative interaction between PHC and Sub on ERC unveils potential cross-area coordination within the brain. The last edge is reversed, possibly attributed to the presence of directed cycles, leading \methodname{} to misjudge the direction. In summary, the results demonstrate the effectiveness of \methodname{} in identifying causal connections between different brain regions on real fMRI datasets.

\section{Conclusion}
In this paper, we study how to acquire multi-granularity causal structure on observational data, which is a practical and significant, but currently under-researched problem. We proposed an effective approach \methodname{} that leverages SAE to extract latent macro-variables, and utilizes MLPs with a simplified acyclicity penalty to speed up the discovery of multi-granularity causal structure. The efficacy of \methodname{} is verified by experiments on synthetic and real-world datasets.

\clearpage

\appendix
\setcounter{table}{0}
\setcounter{equation}{0}
\renewcommand{\thetable}{S\arabic{table}}
\renewcommand{\theequation}{S\arabic{equation}}

\twocolumn[
\begin{@twocolumnfalse}
	\section*{\centering{Multi-granularity Causal Structure Learning\\ Supplementary File \\[25pt]}}
\end{@twocolumnfalse}
]

\section{Algorithm Table}

\begin{algorithm}[!b]
\caption{\methodname{}: Multi-granularity causal structure learning} 
\label{alg:MgCSL}
\textbf{Input}: Observational data $\mathbf{X}$, parameter: $\alpha_1$, $\alpha_2$, $\mu$, $\gamma$, $\eta$, $\rho$\\
\textbf{Output}: Micro-level weighted adjacency matrix $\mathbf{C}$ or multi-granularity weighted adjacency matrix $\mathbf{S}$ with $\mathbf{A}$\\
\begin{algorithmic}[1] 
\STATE Initial parameters $\mathbf{W}_\text{encoder}$, $\mathbf{W}_\text{decoder}$ and $\mathbf{W}_\text{MLP}$
\REPEAT
\STATE Calculate reconstructed data $\mathbf{Y}$ and obtain macro-representations $\mathbf{Z}$ with $\mathbf{X}$ through encoder and decoder via Eq. (2)
\STATE  Reconstruct $\mathbf{\hat{\mathbf{X}}}$ with $\mathbf{X}$ and $\mathbf{Z}$ via Eq. (5)
\STATE Calculate objective function $\mathcal{L}$ via Eq. (10) and optimize Eq. (16) via L-BFGS-B algorithm to search optimal DAG
\STATE Update parameters $\mu$ and $\gamma$ via Eqs. (17) and (18)
\UNTIL{reach convergence or maximal iterations}
\IF{micro-level graph is required}
\STATE \textbf{return} $\mathbf{C}$
\ELSIF{multi-granularity graph is required}
\STATE \textbf{return} $\mathbf{S}$, $\mathbf{A}$
\ENDIF
\end{algorithmic}
\end{algorithm}

The comprehensive procedure of \methodname{} is delineated in Algorithm \ref{alg:MgCSL}. Particularly, line 1 initials parameters of SAE and MLP. Line 2-7 converts the problem of multi-granularity causal graph search to optimize an objective function with acyclicity constraint. Specifically, line 3 leverages SAE to extract causal abstractions from micro-variables, yielding macro-variable representative data $\mathbf{Z}$  and reconstructed data $\mathbf{Y}$. Line 4 fits causal functions with MLPs and obtain reconstructed data $\hat{\mathbf{X}}$ to search potential parent variables.
Line 5 derives the objective function for the task and then solves the objective function minimization problem using the L-BFGS-B algorithm. Line 6 updates the penalty coefficient $\mu$ and Lagrange multiplier $\gamma$, along with the acyclicity constraint. When reaching the convergence or maximal iterations, \methodname{} will stop the iteration in Line 7. Line 8-12 return micro-level DAG or multi-granularity DAG based on the requirement as found causal structure.

\section{Proof of Proposition 1}

\begin{proposition}
\label{eigval}
A WAM $\mathbf{C}\in\mathbb{R}^{d\times d}$ is a DAG if and only if 
\begin{equation}
   \forall i\in[1,d], \lambda_i=0
\end{equation}
where $\lambda_i$ represents the eigenvalues of $\mathbf{D}$.
\end{proposition}

\subsubsection{Proof.}
For $\mathbf{D}$$=$$\mathbf{C}\circ\mathbf{C}$, when there exists a directed cycle of length $s$ in $\mathbf{D}$, then $\text{tr}(\mathbf{D}^s) > 0$. Since the longest directed cycle in $\mathbf{D}\in\mathbb{R}^{d \times d}$ is of length $d$, when there are no directed cycles in $\mathbf{D}$, the following equations hold: 
\begin{equation}
    \forall s\in[1,d], \text{tr}(\mathbf{D}^s) = 0
    \label{traceD}
\end{equation}
For the eigenvalues $\lambda_i(i$$\in$$[1, d])$ of matrix $\mathbf{D}$, we have $\text{tr}(\mathbf{D}) = \sum_{i=1}^d \lambda_i$. In addition, the eigenvalues of $\mathbf{D}^d$ are $\lambda_i^d(i$$\in$$[1, d])$. This implies that  Eq. (\ref{traceD}) can be transformed into:
\begin{equation}
    \forall s\in[1,d], {\sum}_{i=1}^d \lambda_i^s=0
    \label{eigsum}
\end{equation}
Then it can be simplified as:
\begin{equation}
    \forall s\in[1,d], \lambda_i^s=0
    \label{eig}
\end{equation}
(\ref{eigsum})$\Rightarrow$(\ref{eig}):
We prove the statement by contradiction. Suppose that there exists $s$$\in$$[1,d]$ and $j$$\in$$[1,d]$, such that $\lambda_j^s$$\neq$$0$ and $\sum_{i=1}^d \lambda_i^s$$=$$0$ is satisfied. 
(i) If $s$ is an odd number, then $\lambda_i^{s-1}$ (or $\lambda_i^{s+1}$) are non-negative, thus $\sum_{i=1}^d\lambda_i^{s-1}$$\neq$$0$ (or $\sum_{i=1}^d\lambda_i^{s+1}$$\neq$$0$), the original assumption is false. 
(ii) If $s$ is an even number, then $\forall i\in[1,d], \lambda_i^s$$\geq$$0$, thus $\sum_{i=1}^d\lambda_i^s$ cannot be satisfied, the original assumption is false.
Thus, we conclude that (\ref{eigsum})$\Rightarrow$(\ref{eig}) can be satisfied.\\
(\ref{eig})$\Rightarrow$(\ref{eigsum}):
It is obvious that if $\lambda_i^s$$=$$0$, then ${\sum}_{i=1}^d \lambda_i^s$$=$$0$.

Furthermore, if $\lambda_i$$=$$0$, then $\lambda_i^s$$=$$0(s$$\in$$[1,d])$, thus the problem can be finally simplified as follows:
\begin{equation}
   \forall i\in[1,d], \lambda_i=0
\end{equation}

\section{Experiments}

\subsection{Data Simulation}

Given the graph $\mathcal{G}$, we simulate the SEM $\mathbf{x}_i$$=$$f_i(\mathbf{x}_{pa(i)})+N_i$ for all $i$$=$$1,2,\cdots,d$ in the topological order induced by $\mathcal{G}$. We consider the following instances of $f_i$:

$\textbf{Gaussian processes}$: $f_i$ is drawn from Gaussian process applying RBF kernel with one length scale.

$\textbf{Additive Gaussian processes}$: $f_i(\mathbf{x}_{pa(i)})$$=$$\sum_{j\in pa(i)}f_{ji}(\mathbf{x}_j)$, where each $f_{ji}$ is drawn from Gaussian process applying RBF kernel with one length scale.

To decompose macro-variables into micro-level representations, we employ a randomly intialized MLP with one hidden layer of 100 units and sigmoid activation function.

\subsection{Runtime Environments}

We conduct experiments of all methods on a single GPU (NVIDIA RTX 3090, 24 GB memory) with four 2.1 GHz Intel Xeon CPU cores and 503 GB memory.

\subsection{Parameter Configuration}

\begin{table}[!b]
\centering
\scriptsize
\begin{tabular}{|l|l|r|l}
\cline{1-3}
 & parameter &value       &  \\ \cline{1-3}
PC                         & \multicolumn{1}{l|}{significance level $\alpha$}          & 0.05  &  \\ \cline{1-3}
\multirow{5}{*}{DAG-GNN}                & \multicolumn{1}{l|}{learning rate} & 0.003 &  \\ \cline{2-3}   
                                & \multicolumn{1}{l|}{VAE layers}          & [1,64,1,64,1]  &  \\ \cline{2-3}
                                & \multicolumn{1}{l|}{initial $\lambda$}          & 0  &  \\ \cline{2-3}
                                        & \multicolumn{1}{l|}{initial $c$}          &  1 &  \\ \cline{1-3}
\multirow{7}{*}{GAE}                         & \multicolumn{1}{l|}{learning rate} & 0.001 &  \\ \cline{2-3}
                                & \multicolumn{1}{l|}{input units}          &  1 &  \\ \cline{2-3}
                                 & \multicolumn{1}{l|}{hidden layers}          &  1 &  \\ \cline{2-3}
                                 & \multicolumn{1}{l|}{hidden units}          &  4 &  \\ \cline{2-3}
                                & \multicolumn{1}{l|}{$\lambda$}          &  0 &  \\ \cline{2-3}
                                & \multicolumn{1}{l|}{initial $\alpha$}          &  0 &  \\ \cline{2-3}
                                & \multicolumn{1}{l|}{initial $\rho$}          &  1 &  \\ \cline{1-3}
\multirow{6}{*}{GraN-DAG}    & \multicolumn{1}{l|}{learning rate} & 0.001 &  \\ \cline{2-3}
                             & \multicolumn{1}{l|}{hidden layers} & 2     &  \\ \cline{2-3}
                             & \multicolumn{1}{l|}{hidden units}  & 10    &  \\ \cline{2-3}
                             & \multicolumn{1}{l|}{PNS threshold} & 0.75  &  \\ \cline{2-3}
                             & \multicolumn{1}{l|}{initial $\lambda$}   & 0     &  \\ \cline{2-3}
                             & \multicolumn{1}{l|}{initial $\mu$}       & 0.001 &  \\ \cline{1-3}
\multirow{5}{*}{NOTEARS-MLP} & \multicolumn{1}{l|}{hidden units}  & 10    &  \\ \cline{2-3}
                             & \multicolumn{1}{l|}{hidden layers} & 1     &  \\ \cline{2-3}
                             & \multicolumn{1}{l|}{$\lambda$}    & 0.01     &  \\ \cline{2-3}
                             & \multicolumn{1}{l|}{initial $\alpha$}    & 0     &  \\ \cline{2-3}
                             & \multicolumn{1}{l|}{initial $\rho$}      & 1     &  \\ \cline{1-3}
\multirow{11}{*}{\methodname{}}    & \multicolumn{1}{l|}{$\alpha_1$}  &   0.1  &  \\ \cline{2-3}
                            & \multicolumn{1}{l|}{$\alpha_2$}  &   0.01  &  \\ \cline{2-3}
                            & \multicolumn{1}{l|}{$\eta$}  &  300   &  \\ \cline{2-3}
                            & \multicolumn{1}{l|}{$\rho$}  &  0.25   &  \\ \cline{2-3}
                            & \multicolumn{1}{l|}{initial $\gamma$}  &  0   &  \\ \cline{2-3}
                            & \multicolumn{1}{l|}{initial $\mu$}  &  0.001   &  \\ \cline{2-3}
                            & \multicolumn{1}{l|}{tolerance of $\mathcal{H}$}  &  0.1   &  \\ \cline{2-3}
                            & \multicolumn{1}{l|}{maximum of $\mu$}  &  $10^{16}$   &  \\ \cline{2-3}
                            & \multicolumn{1}{l|}{$\epsilon$}  &   0.2  &  \\ \cline{2-3}
                                & \multicolumn{1}{l|}{SAE layers}          &  [$d$,0.75$d$,5,0.75$d$,$d$] &  \\ \cline{2-3}
                                & \multicolumn{1}{l|}{MLP layers}          &  [$d$$+$5,10,1] &  \\ \cline{1-3}
\end{tabular}
\caption{Parameter configuration of methods}
\label{parameter}
\end{table}

We present the parameters we use of all methods in Table \ref{parameter}, which are either optimized as suggested in the respective literature or given in shared codes. We also present the parameters of our proposed \methodname{} not mentioned in the main text. Specifically, we tried $\eta$$\in$\{10, 30, 100, 300, 500\} and $\rho$$\in$\{0.1, 0.25, 0.5, 0.75\}. A larger $\eta$ and a smaller $\rho$ enable faster convergence, but may lead to a low accuracy due to jumping out of the iteration too early, so we chose $\eta$=$300$ and $\rho$=$0.25$ to manage faster convergence and accuracy. We separately initialize $\gamma$ and $\mu$ to $0$ and 0.001, and set tolerance of $\mathcal{H}$, maximum of $\mu$ to 0.1, $10^{16}$ to determine the termination of the optimization process. 

\subsection{Supplementary Experiments}

\begin{table*}[htbp]
\tiny
\setlength\tabcolsep{3pt} 
\centering
\begin{tabular}{c l| r r r r| r|| r r r r| r}
\hline
      & &\multicolumn{5}{c|}{\emph{Erd\H os-R\'enyi} graph} &\multicolumn{5}{c}{\emph{Scale-Free} graph}\\
\hline
$\#$Variables && Precision(\%)$\uparrow$ & Recall(\%)$\uparrow$ & F1(\%)$\uparrow$ & SHD$\downarrow$ & Runtime(s)$\downarrow$ &  Precision(\%)$\uparrow$ & Recall(\%)$\uparrow$ & F1(\%)$\uparrow$ & SHD$\downarrow$ & Runtime(s)$\downarrow$ \\
\hline
\multicolumn{1}{c|}{\multirow{7}{*}{20}}
      & PC & 40.09$\pm$8.73 & 30.25$\pm$8.93 & 34.40$\pm$8.88 & 34.50$\pm$5.44 & 0.24$\pm$0.04 & 36.50$\pm$10.77 & 31.35$\pm$10.13 & 33.62$\pm$10.25 & 35.10$\pm$5.65 & 0.36$\pm$0.13  \\
\multicolumn{1}{c|}{} & GES & 46.72$\pm$8.78 & 31.75$\pm$6.35 & 37.73$\pm$7.16 & 30.70$\pm$3.13 & 8.74$\pm$1.49 & 42.65$\pm$7.85 & 35.95$\pm$7.65 & 38.90$\pm$7.37 & 31.50$\pm$2.88 & 9.87$\pm$1.06  \\
      \cline{2-12}
\multicolumn{1}{c|}{} & DAG-GNN & 89.85$\pm$10.63 & 15.50$\pm$5.87 & 26.05$\pm$8.72 & 34.00$\pm$2.36 & 2996.17$\pm$345.26 & 74.79$\pm$14.63 & 11.08$\pm$4.12 & 18.96$\pm$6.27 & 33.00$\pm$1.56 & 2556.91$\pm$667.25  \\
\multicolumn{1}{c|}{} & GAE & 76.09$\pm$17.46 & 26.75$\pm$12.42 & 37.33$\pm$14.05 & 31.60$\pm$4.27 & 1328.48$\pm$92.95  & 72.04$\pm$27.17 & 25.14$\pm$6.38 & 35.35$\pm$7.19 & 32.10$\pm$7.29 & 9484.15$\pm$459.68  \\
\multicolumn{1}{c|}{} & GraN-DAG    & 81.18$\pm$21.71 & 20.50$\pm$5.37 & 31.58$\pm$5.09 & 34.90$\pm$3.84 & 306.18$\pm$14.17 & 77.37$\pm$18.78 & 39.19$\pm$12.95 & 49.10$\pm$9.34 & 27.70$\pm$2.75 & 354.22$\pm$1.87\\
\multicolumn{1}{c|}{}  & NOTEARS-MLP & 93.56$\pm$5.14 & 42.25$\pm$9.61 & 57.76$\pm$9.72 & 23.80$\pm$4.26 & 16.51$\pm$3.70 & 79.58$\pm$11.30 & 48.65$\pm$8.64 & 60.07$\pm$9.22 & 21.40$\pm$4.93 & 21.96$\pm$4.23  \\
\multicolumn{1}{c|}{}  & \methodname{} & \bf 96.92$\pm$4.33 & \bf 59.00$\pm$11.56 & \bf 72.85$\pm$9.63 & \bf 16.90$\pm$4.72 & 5.51$\pm$1.59  & \bf 96.42$\pm$3.75 & \bf 57.30$\pm$8.99 & \bf 71.35$\pm$6.70 & \bf 16.50$\pm$2.92 & 4.12$\pm$1.35  \\
\hline
\multicolumn{1}{c|}{\multirow{7}{*}{50}}
      & PC & 30.18$\pm$4.08 & 27.90$\pm$2.60 & 28.95$\pm$3.11 & 109.90$\pm$8.61 & 1.52$\pm$0.69 & 25.38$\pm$4.91 & 24.12$\pm$4.24 & 24.72$\pm$4.49 & 118.30$\pm$8.07 & 5.52$\pm$5.25  \\
\multicolumn{1}{c|}{}  & GES & 45.96$\pm$4.03 & 34.50$\pm$4.14 & 39.34$\pm$3.78 & 85.60$\pm$4.35 & 402.53$\pm$57.24 & 38.90$\pm$4.92 & 32.58$\pm$4.79 & 35.42$\pm$4.75 & 97.70$\pm$7.10 & 426.46$\pm$31.33  \\
      \cline{2-12}
\multicolumn{1}{c|}{} & DAG-GNN & \bf 92.04$\pm$5.83 & 13.50$\pm$6.28 & 23.06$\pm$9.27 & 86.60$\pm$6.29 & 2948.64$\pm$332.51 & 91.54$\pm$6.77 & 15.26$\pm$5.87 & 25.69$\pm$8.77 & 82.40$\pm$5.56 & 3464.21$\pm$384.46  \\
\multicolumn{1}{c|}{} & GAE & 78.11$\pm$23.26 & 12.70$\pm$6.31 & 20.73$\pm$8.93 & 92.60$\pm$7.52 & 4348.12$\pm$3636.96  & 78.41$\pm$21.80 & 12.99$\pm$5.94 & 21.15$\pm$7.23 & 89.70$\pm$5.62 & 4680.90$\pm$2141.31  \\
\multicolumn{1}{c|}{} & GraN-DAG  & 13.35$\pm$2.38 & 19.90$\pm$6.49 & 15.49$\pm$2.53 & 207.00$\pm$40.21 & 851.68$\pm$165.55 & 14.61$\pm$3.04 & 24.02$\pm$8.54 & 17.53$\pm$3.87 & 212.20$\pm$64.60 & 634.74$\pm$17.62\\
\multicolumn{1}{c|}{}  & NOTEARS-MLP & 90.58$\pm$2.84 & 43.40$\pm$4.55 & 58.57$\pm$4.43 & 58.30$\pm$5.25 & 254.20$\pm$44.55  & 80.37$\pm$5.93 & 40.83$\pm$4.16 & 54.03$\pm$4.22 & 63.70$\pm$4.22 & 166.14$\pm$32.26  \\
\multicolumn{1}{c|}{} & \methodname{} & \bf 92.07$\pm$6.24 & \bf 51.10$\pm$8.67 & \bf 65.12$\pm$6.76 & \bf 53.70$\pm$6.98 & 16.92$\pm$4.26  & \bf 94.67$\pm$2.78 & \bf 45.57$\pm$9.20 & \bf 61.03$\pm$8.54 & \bf 55.10$\pm$8.70 & 22.81$\pm$3.24  \\
\hline
\multicolumn{1}{c|}{\multirow{7}{*}{100}}
      & PC & 20.42$\pm$2.68 & 25.30$\pm$3.58 & 22.59$\pm$3.03 & 284.60$\pm$10.96 & 7.97$\pm$1.17 & 19.59$\pm$2.61 & 24.26$\pm$3.27 & 21.67$\pm$2.86 & 291.50$\pm$15.11 & 14.49$\pm$3.66  \\
\multicolumn{1}{c|}{} & GES & 39.46$\pm$4.49 & 34.88$\pm$4.20 & 36.97$\pm$3.99 & 194.75$\pm$9.42 & 5485.93$\pm$618.06 & 35.20$\pm$4.49 & 33.16$\pm$4.54 & 34.13$\pm$4.44 & 215.00$\pm$14.20 & 5854.72$\pm$638.99  \\
      \cline{2-12}
\multicolumn{1}{c|}{} & DAG-GNN & \bf 92.82$\pm$4.43 & 9.45$\pm$2.40 & 17.08$\pm$3.99 & 181.20$\pm$4.69 & 1579.77$\pm$102.75 & 85.73$\pm$3.43 & 11.22$\pm$1.39 & 19.80$\pm$2.09 & 175.10$\pm$2.64 & 3845.05$\pm$536.62  \\
\multicolumn{1}{c|}{} & GAE & 66.90$\pm$25.78 & 10.80$\pm$6.53 & 16.82$\pm$8.31 & 197.70$\pm$17.20 & 10329.01$\pm$313.16  & 78.80$\pm$27.42 & 8.58$\pm$6.71 & 14.19$\pm$9.22 & 190.50$\pm$14.28 & 10310.63$\pm$221.60  \\
\multicolumn{1}{c|}{}  & GraN-DAG    & 0.67$\pm$0.56 & 2.05$\pm$2.25 & 0.98$\pm$0.90 & 751.90$\pm$226.26 & 2583.10$\pm$206.32 & 1.50$\pm$0.80 & 4.57$\pm$3.36 & 2.21$\pm$1.28 & 725.70$\pm$162.48 &1740.89$\pm$311.92\\
\multicolumn{1}{c|}{} & NOTEARS-MLP & 90.87$\pm$2.68 & \bf 34.95$\pm$4.57 & \bf 50.36$\pm$5.11 & \bf 132.70$\pm$9.06 & 1832.14$\pm$282.42  & 79.74$\pm$4.39 & 34.72$\pm$2.76 & 48.34$\pm$3.21 & 136.50$\pm$5.58 & 1061.95$\pm$97.38  \\
\multicolumn{1}{c|}{}  & \methodname{} &  90.99$\pm$6.10 & 34.30$\pm$3.97 & 49.59$\pm$3.89 & 137.70$\pm$8.42 & 163.32$\pm$35.50  & \bf 89.92$\pm$4.78 & \bf 36.19$\pm$5.74 & \bf 51.23$\pm$5.43 & \bf 132.10$\pm$8.29 & 193.30$\pm$23.79  \\
\hline
\end{tabular}
\caption{Results on \emph{nonlinear models with Gaussian processes}.}
\label{syntgp}
\end{table*}

We show the results on nonlinear models with Gaussian processes in Table \ref{syntgp}. Overall \methodname{} achieves better or comparable performance across all datasets. Also we observe that \methodname{} takes less time than most baselines, affirming its efficiency.

\end{document}